\documentclass{article} % For LaTeX2e
\usepackage{iclr2023_conference,times}

\usepackage{amsmath,amsfonts,bm}

\def\eqref#1{equation~\ref{#1}}
\def\1{\bm{1}}

\DeclareMathAlphabet{\mathsfit}{\encodingdefault}{\sfdefault}{m}{sl}
\SetMathAlphabet{\mathsfit}{bold}{\encodingdefault}{\sfdefault}{bx}{n}

\usepackage{hyperref}
\usepackage{url}

\usepackage{booktabs}
\usepackage{mathtools}
\usepackage{amsthm}
\usepackage{subfigure}
\usepackage{tabularx}
\usepackage{diagbox}
\usepackage{enumitem}
\usepackage{graphicx}

\theoremstyle{plain}
\newtheorem{theorem}{Theorem}[section]

\theoremstyle{definition}
\newtheorem{definition}[theorem]{Definition}

\theoremstyle{remark}

\title{From ChebNet to ChebGibbsNet}

\author{Jie Zhang, Min-Te Sun\\
Department of Computer Science and Information Engineering\\
National Central University\\
Taiwan\\
\texttt{hazdzz@g.ncu.edu.tw, msun@csie.ncu.edu.tw}
}

\iclrfinalcopy % Uncomment for camera-ready version, but NOT for submission.
\begin{document}

\maketitle

\begin{abstract}
Recent advancements in Spectral Graph Convolutional Networks (SpecGCNs) have led to state-of-the-art performance in various graph representation learning tasks. To exploit the potential of SpecGCNs, we analyze corresponding graph filters via polynomial interpolation, the cornerstone of graph signal processing. Different polynomial bases, such as Bernstein, Chebyshev, and monomial basis, have various convergence rates that will affect the error in polynomial interpolation. Although adopting Chebyshev basis for interpolation can minimize maximum error, the performance of ChebNet is still weaker than GPR-GNN and BernNet. \textbf{We point out it is caused by the Gibbs phenomenon, which occurs when the graph frequency response function approximates the target function.} It reduces the approximation ability of a truncated polynomial interpolation. In order to mitigate the Gibbs phenomenon, we propose to add the Gibbs damping factor with each term of Chebyshev polynomials on ChebNet. As a result, our lightweight approach leads to a significant performance boost. Afterwards, we reorganize ChebNet via decoupling feature propagation and transformation. We name this variant as \textbf{ChebGibbsNet}. Our experiments indicate that ChebGibbsNet is superior to other advanced SpecGCNs, such as GPR-GNN and BernNet, in both homogeneous graphs and heterogeneous graphs.
\end{abstract}

\section{Introduction}
High dimensional data are ordinarily represented as graphs in a variety of areas, such as citation, energy, sensor, social, and traffic networks. Consequently, Graph Neural Networks (GNNs), also known as CNNs on graphs, have shown a tremendous promise. 

Since the emergence of GCN~\citep{DBLP:conf/iclr/KipfW17}, numerous GNNs have been developed to generalize convolution operations on graphs. Graph convolution is based on graph signal processing~\citep{6494675,SHUMAN2016260}, where the graph filter is a crucial component. A graph filter is a matrix which processes a graph signal by amplifying or attenuating its corresponding graph Fourier coefficients. Generally, a graph filter is a combination of different powers of a graph shift operator, which is a normalized adjacency matrix or Laplacian matrix.

Recently, enlightened by graph diffusion~\citep{vigna_2016,MASUDA20171}, GNNs such as SGC~\citep{pmlr-v97-wu19e}, APPNP~\citep{klicpera2018combining}, S$^{2}$GC~\citep{zhu2021simple}, GPR-GNN~\citep{chien2021adaptive}, and BernNet~\citep{he2021bernnet} focus on designing graph filters, and demonstrate strong performance in node classification. We call those GNNs as Spectral Graph Convolutional Networks (SpecGCNs). In order to exploit the potential of SpecGCNs, we summarize those SpecGCNs into a single architecture. We find most of those SpecGCNs' graph filters are monomials basis, with the exception of ChebNet~\citep{DBLP:conf/nips/DefferrardBV16} and BernNet. Besides, except GPR-GNN and BernNet, other SpecGCNs' graph filters are accompanied with fixed coefficients. 

Next, we point out that the Gibbs phenomenon~\citep{Hewitt1979,Jerri1998}, which troubles signal processing, is also a tricky issue for SpecGCNs. When the target graph frequency response function is discontinuous or singular at some points in the polynomial interpolation interval, the Gibbs phenomenon will occur around discontinuities or singularities. Consequently, polynomial-based graph frequency response function will dramatically oscillate near discontinuities or singularities. Those oscillations are called Gibbs oscillations. For graph signal processing that approximates via truncated polynomials, Gibbs oscillations reduce approximation accuracy.

In this paper, we apply the Gibbs damping factor with each term of Chebyshev polynomials of the first kind (hereinafter referred to as Chebyshev polynomials). The Gibbs damping factor is a family of factors that are committed to mitigate Gibbs oscillations. We test ChebNet with various Gibbs damping factors for semi-supervised and supervised node classification on both homogeneous and heterogeneous graphs. The experimental results exhibit that ChebyNet with several Gibbs damping factors indeed can refine performance in the majority of datasets.

Similar to APPNP, we reorganize ChebNet with Gibbs damping factors via decoupling feature propagation and transformation. Besides, in each term, we add a learnable coefficient to represent corresponding Chebyshev coefficient. We term this model as ChebGibbsNet. Compare with state-of-the-art SpecGCNs, ChebGibbsNet demonstrates powerful performance. 

The contributions in this paper are as follows:
\begin{enumerate}[label=\roman*]
  \item We indicate that the Gibbs phenomenon troubles graph signal processing.
  \item We apply Chebyshev polynomials with various Gibbs damping factors as the graph filter to propose our SpecGCN named ChebGibbsNet.
  \item We conduct extensive experiments on a variety of homogeneous and heterogeneous datasets to validate the performance of ChebGibbsNet.
\end{enumerate}

% The rest of the paper is organized as follows. Section 2 provides the related works of this paper. Section 3 outlines the necessary background knowledge. Section 4 describes the proposed method. Second 5 presents and analyze the experiment results, and the conclusion and future work are given in Section 6.
\section{Related Work}\label{sec:rw}
\noindent\textbf{Spectral Graph Convolutional Networks.}
Based on graph signal processing, ChebNet~\citep{DBLP:conf/nips/DefferrardBV16} is proposed with a localized graph filter. ChebNet is the first SpecGCN with graph filter modification, which utilizes Chebyshev polynomials. Then GCN~\citep{DBLP:conf/iclr/KipfW17} improves ChebyNet by proposing a method called the renormalization trick which inspires several SpecGCNs to design filters. As a simplified version of multi-layer GCN, SGC~\citep{pmlr-v97-wu19e} eliminates nonlinear activation functions and Dropout method~\citep{srivastava14a} in order to retain performance and achieve the same results as GCN.

Since graph convolution is related to graph signal processing, several researchers have tried to design SpecGCNs from the view of devising graph filters. One common approach is focusing on graph diffusion~\citep{vigna_2016,MASUDA20171}. The Personalized PageRank filter~\citep{10.1145/775152.775191} is a well-known graph diffusion filter adopted by APPNP~\citep{klicpera2018combining}. Subsequently, S$^{2}$GC~\citep{zhu2021simple} adopts Markov Diffusion kernel~\citep{4053117} to demonstrate its graph performance in the node classification task. Afterwards, based on the Generalized PageRank filter~\citep{10.1145/1148170.1148225,doi:10.1137/140976649} with learnable coefficients, GPR-GNN~\citep{chien2021adaptive} shows its high performance in both homogeneous graphs and heterogeneous graphs. Hereafter, BernNet~\citep{he2021bernnet} utilizes Bernstein polynomials to adaptively learn any graph filter.

\noindent\textbf{Graph Signal Processing and Polynomial Interpolation.}
As a branch of signal processing, the graph signal processing is based on polynomial interpolation, a mathematical approach that utilizes a finite polynomial to approximate any target function. In traditional signal processing, a polynomial-based function interpolates nodes, which are sampled from an interval, to approximate target signal function. As for graph signal processing, the corresponding sampling nodes are eigenvalues of a graph shift operator. Due to high time complexity of eigendecomposition, directly obtaining eigenvalues would not be generally executed. Hence, how to design an appropriate graph filter is a challenge for SpecGCNs.

\noindent\textbf{Kernel Polynomial Method and Gibbs Damping Factors}
In some areas of physics, such as thermodynamics and quantum mechanics, the study of the eigenfunctions of a dynamical matrix or Hamiltonian operator is crucial. Based on polynomial interpolation, kernel polynomial method is invented as a core component for above studies~\citep{RevModPhys.78.275}. The approach called Chebyshev expansion with modified moments, i.e., Chebyshev polynomials with Gibbs damping factors, is introduced to mitigate Gibbs oscillations while kernel polynomial method is widely applied in physics. 
\section{Preliminary and Background}\label{sec:pre}
\subsection{Homogeneous Graphs and Heterogeneous Graphs}
A undirected graph $G$ is represented as $G = \{\mathcal{V},\mathcal{E}\}$, where $\mathcal{V} = \{v_{0},v_{1},...,v_{n-1}\}$ is the set of vertices with $|\mathcal{V}| = n$, and $\mathcal{E} \subseteq \mathcal{V}\times\mathcal{V}$ is the set of edges. Let $\mathbf{A} \in \mathbb{R}^{n \times n}$ denote the adjacency matrix of $G$. Given two nodes $u$ and $v$, $\mathbf{A}(u,v) = 1$ if there is an edge between node $u$ and node $v$. Otherwise, $\mathbf{A}(u,v) = 0$. The diagonal degree matrix $\mathbf{D}\in\mathbb{R}^{{n}\times{n}}$ is obtained by $\mathbf{D}(u,u) = \sum_{v\in\mathcal{V}}\mathbf{A}(u,v)$. Then, the combinatorial Laplacian~\citep{chung1997spectral} is defined as $\mathbf{L} = \mathbf{D} - \mathbf{A}$ for an undirected graph $G$. The symmetric normalized Laplacian matrix is defined as $\mathcal{L} = \mathbf{I}_{n} - \mathbf{D}^{-\frac{1}{2}}\mathbf{A}\mathbf{D}^{-\frac{1}{2}}$. We denote the symmetric normalized adjacency matrix as $\mathcal{A} = \mathbf{D}^{-\frac{1}{2}}\mathbf{A}\mathbf{D}^{-\frac{1}{2}}$.

Graphs can be either homogeneous or heterogeneous. The homophily and heterophily of a graph are used to describe the relation of labels among nodes. A homogeneous graph is a graph where the labels of all the nodes are consistent. On the contrary, in a heterogeneous graph, labels for nodes are of different types. The node homophily index for a graph~\citep{Pei2020Geom-GCN} is denoted as 
\begin{definition}[Node Homophily]
\begin{equation}
\centering
\begin{split}
\mathcal{H}_{\mathrm{node}}(G) = \frac{1}{|\mathcal{V}|}\sum_{u \in \mathcal{V}}{\frac{\left|\{v|v\in\mathcal{N}_{u},\mathcal{Y}_{v}=\mathcal{Y}_{u}\}\right|}{|\mathcal{N}_{u}|}},   
\end{split}
\end{equation}
where $\mathcal{N}_{u}$ is the neighbor set of node $u$ and $\mathcal{Y}_{u}$ is the label of node $u$.
\end{definition}
Note that $\mathcal{H}_{\mathrm{node}}(G) \rightarrow 1$ indicates strong homophily and vice versa. 

\subsection{Graph Signal Processing}
A column feature vector $\mathbf{x}$ in graph signal processing (GSP) is called a graph signal~\citep{6409473,6808520}. A graph shift operator (GSO)~\citep{6409473,6808520} is a matrix which defines how a graph signal is shifted from one node to its neighbors based on the graph topology. More specifically, GSO is a local operator that replaces graph signal value of each node with linear combination of its neighbors'. In graph signal processing, it is common to take a normalized adjacency matrix or Laplacian matrix as a GSO.

A function $h(\cdot)$ of an eigenvalue $\lambda_{j}$, where $j\in[0, n-1]$, as $h(\lambda_{j})$ is called the graph frequency response function. It is a discrete function, which extends the convolution theorem from digital signal processing to graphs. For a normalized Laplacian matrix, its graph frequency response function is defined as $h(\lambda_{j}) = \lambda_{j}$. For a normalized adjacency matrix, its graph frequency response function is $h(\lambda_{j}) = 1 - \lambda_{j}$. We denote $g(\lambda_{j})$ as graph shifting for an eigenvalue $\lambda_{j}$.

A graph filter $\mathbf{H}(\mathbf{S}) \in \mathbb{C}^{n \times n}$~\citep{6409473,6808520} is a function of a graph shift operator $\mathbf{S}$. Apparently, a GSO is a graph filter. In graph signal processing, it is common to utilize a polynomial-based graph filter which is defined as $\mathbf{H}(\mathbf{S}) = \sum^{K}_{k=0}{{\zeta}_{k}\mathbf{S}^{k}}$, where ${\zeta}_{k}$ is the corresponding coefficient. This kind of graph filter is named Moving-Average (MA) filter~\citep{7581108}, which is also named Finite Impulse Response (FIR) filter. The capacity of a SpecGCN with an FIR filter is determined by the degree $k$ of a polynomial. We denote it as a MA$_{K}$ or FIR$_{K}$ filter.

For an undirected graph $G$, its graph filter can be eigendecomposed as $\mathbf{H}(\mathbf{S}) = \bf{U}\bf{\Lambda}\mathbf{U}^{*}$, where $\mathbf{U} \in \mathbb{R}^{n \times n}$ is a matrix of orthogonal eigenvectors, $\mathbf{\Lambda} = \mathrm{diag}\left([h(g(\lambda_{0})),...,h(g(\lambda_{n-1}))]\right) \in \mathbb{R}^{n \times n}$ is a diagonal matrix of filtered eigenvalues, and ${*}$ means conjugate transpose. Since $\mathbf{U}$ is real-valued, we have $\mathbf{U}^{*} = \mathbf{U}^{\top}$.

Based on the theory of graph signal processing, the graph Fourier transform for a graph signal vector $\mathbf{x}$ on an undirected graph is defined as $\hat{\mathbf{x}} = \mathbf{U}^{*}\mathbf{x}$, and the inverse graph Fourier transform is $\mathbf{x} = \mathbf{U}\hat{\mathbf{x}}$. Given a graph filter $\mathbf{H}(\mathbf{S})$ and a feature matrix $\mathbf{X}$, the convolution operator on a graph is defined as $\mathbf{H}(\mathbf{S})\ast_{G}\mathbf{X} = \mathbf{U}\left((\mathbf{U}^{*}\mathbf{H}(\mathbf{S}))\odot(\mathbf{U}^{*}\mathbf{X})\right)=\mathbf{U}{h({g(\mathbf{\Lambda})})}\mathbf{U}^{*}\mathbf{X}=\sum_{i=0}^{n-1}{h(g(\lambda_{i}))\mathbf{u}_{i}{\mathbf{u}_{i}}^{*}}$, where ${h({g(\mathbf{\Lambda})})}$ is a matrix form of the graph frequency response function, and $\mathbf{u}_{0},\mathbf{u}_{1},...,\mathbf{u}_{n-1}$ are eigenvectors of $\mathbf{U}$.

\subsection{From ChebNet to GCN}
ChebNet is the first SpecGCN with a localized graph filter based on Chebyshev polynomials. Through three-term recurrence relations, Chebyshev polynomials can be obtained as follows.
\begin{definition}[Chebyshev Polynomials of the first kind]
\begin{equation}
\centering
\begin{split}
T_{k}(x) = \left\{
\begin{aligned}
& 1 \quad & \text{if}\ {k=0},\\
& x \quad & \text{if}\ {k=1},\\
& 2x\cdot{T}_{k-1}(x) - T_{k-2}(x) & \text{if}\ {k \ge 2},
\end{aligned}
\right.
\end{split}
\end{equation}
where $x \in [-1, 1]$.
\end{definition}
Since eigenvalues of a normalized Laplacian $\mathcal{L}$ are in $[0, 2]$, consider the orthogonality of Chebyshev polynomials, it is forbidden to directly replace $x$ by $\mathcal{L}$. Hence, the scaled normalized Laplacian $\widetilde{\mathcal{L}} = \frac{2}{\lambda_{\max}}\mathcal{L} - \mathbf{I}_{n}$ is proposed, where $\lambda_{\max}$ denotes the maximized eigenvalue of the normalized Laplacian. In the original definition of ChebNet, a graph convolution layer is defined as $\mathbf{H}(\widetilde{\mathcal{L}})\mathbf{X} = \sum_{k=0}^{K}c_{k}T(\widetilde{\mathcal{L}})\mathbf{X}$, where $c_{k}$ denotes the $k$-th Chebyshev coefficient. In practice, the graph convolution layer is defined as 
\begin{equation}
\centering
\begin{split}
\mathbf{Z}^{(l+1)} &= \sigma(\sum_{k=0}^{K}{T}_{k}(\widetilde{\mathcal{L}})\mathbf{Z}^{(l)}\mathbf{W}^{(l)}),
\end{split}
\end{equation}
where $\sigma(\cdot)$ is a non-linear activation function such as ReLU~\citep{10.5555/3104322.3104425}, $\mathbf{Z}^{(l)}$ denotes the $l$-th hidden layer, and $\mathbf{Z}^{(0)} = \mathbf{X}$. The Chebyshev coefficient vector is replaced by a learnable weight matrix $\mathbf{W}^{(l)}$. This operation bewilders numerous researches who deem the $k$-th Chebyshev coefficient is fixed as $c_{k} = 1$, and misinterpret $\mathbf{Z}^{(l)}\mathbf{W}^{(l)}$ as feature transformation.

\noindent\textbf{GCN.}
To simplify ChebNet, a linear version named GCN is proposed in~\citep{DBLP:conf/iclr/KipfW17}. Furthermore, the renormalization trick is proposed. It is defined as $\widetilde{\mathcal{A}} = \widetilde{\mathbf{D}}^{-\frac{1}{2}}(\mathbf{A}+\eta\mathbf{I}_{n})\widetilde{\mathbf{D}}^{-\frac{1}{2}}$, where $\widetilde{\mathbf{D}}(u,u) = \sum_{v \in \mathcal{V}}(\mathbf{A}+\eta\mathbf{I}_{n})(u,v)$, and $\eta=1$ in general. Then, a graph convolution layer of GCN is defined as $\mathbf{Z}^{(l+1)} = \sigma(\widetilde{\mathcal{A}}\mathbf{Z}^{(l)}\mathbf{W}^{(l)})$

\section{Proposed Method}
\subsection{Polynomial Interpolation for Spectral Graph Convolutional Networks}
By decoupling feature transformation and propagation, SpecGCNs can be generalized into a single architecture as
\begin{equation}\label{eq:SpecGCN}
\centering
\begin{split}
\hat{\mathbf{Y}}_{\text{SpecGCN}} = \mathrm{Softmax}\left(\mathbf{H}(\mathbf{S})\cdot{f}_{\Theta}(\mathbf{X})\right),
\end{split}
\end{equation}
where $f_{\Theta}(\mathbf{X})$ is an Multi-Layer Perceptron (MLP). In this unified architecture, the effectiveness of the graph filter is accentuated. Then, we summarize the category of the basis and the corresponding coefficients of SpecGCNs we mentioned in Section~\ref{sec:rw} into Table ~\ref{tab:graph_filter_SpecGCNs}. Next, we discuss the process of polynomial interpolation via the graph frequency response function.
\begin{table}[!t]
\centering
\caption{A summary of graph filters for existing SpecGCNs.}
\begin{tabular}{l|cccccccc}
\toprule
&ChebNet &GCN &SGC &APPNP &S$^{2}$GC &GPR-GNN &BernNet \\
\midrule
Basis &Cheb. &Mono. &Mono. &Mono. &Mono. &Mono. &Bern.\\
Coefficients &Learnable &Fixed &Fixed &Fixed &Fixed &Learnable &Learnable\\
\bottomrule
\end{tabular}
\label{tab:graph_filter_SpecGCNs}
\end{table}

\begin{theorem}[Weierstrass Approximation Theorem~\citep{weierstrass_2013}]\label{theo:wat}
Suppose $f$ be a continuous function on $[a, b]$. For every $\epsilon > 0$, there exists a polynomial $p$ such that $\lVert{f(x)-p(x)}\rVert_{\infty}<\epsilon$.
\end{theorem}
Theorem~\ref{theo:wat} tells us that we can utilize any polynomial-based graph frequency response function to approximate the target function. The whole approximation process can be described into two steps. The first step is graph shifting, i.e., $\lambda \rightarrow {g(\lambda)}$. The second step is graph polynomial interpolation, which is formulated as follows.
\begin{definition}[Graph Polynomial Interpolation]
Given $n$ eigenvalues of a normalized Laplacian after graph shifting as $\widetilde{\lambda}_{j} = g(\lambda_{j})$, where $j \in [0, n-1]$, and a $K \in [0, n-1]$ order polynomial-based graph frequency response function $h(\widetilde{\lambda})$ with corresponding coefficients $\zeta_{k}$. For a target graph frequency response function $f(\widetilde{\lambda})$, the graph polynomial interpolation is to solve a Vandermonde linear system as follows.
\begin{equation}\label{eq:ggfa}
\centering
\begin{bmatrix}
1 &\widetilde{\lambda}_{0} &\widetilde{\lambda}_{0}^{2} &\cdots &\widetilde{\lambda}_{0}^{K}\\
1 &\widetilde{\lambda}_{1} &\widetilde{\lambda}_{1}^{2} &\cdots &\widetilde{\lambda}_{1}^{K}\\
\vdots &\vdots &\vdots &\vdots &\vdots\\
1 &\widetilde{\lambda}_{n-1} &\widetilde{\lambda}_{n-1}^{2} &\cdots &\widetilde{\lambda}_{n-1}^{K}
\end{bmatrix}
\begin{bmatrix}
\zeta_{0}\\
\zeta_{1}\\
\vdots\\
\zeta_{K}
\end{bmatrix}
=
\begin{bmatrix}
h(\widetilde{\lambda}_{0})\\
h(\widetilde{\lambda}_{1})\\
\vdots\\
h(\widetilde{\lambda}_{n-1})
\end{bmatrix}
\approx
\begin{bmatrix}
f(\widetilde{\lambda}_{0})\\
f(\widetilde{\lambda}_{1})\\
\vdots\\
f(\widetilde{\lambda}_{n-1})
\end{bmatrix}
\end{equation}
\end{definition}

On the basis of~\citep{Gautschi2012}, the error function can be defined as follows.
\begin{definition}[Graph Polynomial Interpolation Error Function]
Denote $\widetilde{\lambda}_{j} = g(\lambda_{j}) \in [a, b]$ as an eigenvalue of a normalized Laplacian after graph shifting, where $j \in [0, n-1]$. Given a target graph frequency response function $f(\widetilde{\lambda})$ and a graph frequency response function $h(\widetilde{\lambda})$ based on polynomials with $K \in [0, n-1]$ orders. The error function $e(\widetilde{\lambda})$ is defined as
\begin{equation}\label{eq:error}
\centering
\begin{split}
e(\widetilde{\lambda}) = f(\widetilde{\lambda}) - h(\widetilde{\lambda})
=\frac{f^{(n)}\left(\xi(\widetilde{\lambda})\right)}{n!}\Pi_{j=0}^{n-1}(\widetilde{\lambda} - \widetilde{\lambda}_{j}),
\end{split}
\end{equation}
where $\xi(\widetilde{\lambda}) \in (a, b)$ is an arbitrary number.
\end{definition}
In graph signal processing, the target graph frequency response function is unknown. Thus, our goal is to minimize maximum errors as
\begin{equation}\label{eq:minmax}
\centering
\begin{split}
\min_{\widetilde{\lambda}_{0},\widetilde{\lambda}_{1},\cdots,\widetilde{\lambda}_{n-1}}{\max_{\widetilde{\lambda}\in[a,b]}{|\Pi_{j=0}^{n-1}(\widetilde{\lambda} - \widetilde{\lambda}_{j})|}}.
\end{split}
\end{equation}
It can be derived from Eq.~\ref{eq:error} and Eq.~\ref{eq:minmax} that a monomials basis with appropriate eigenvalues is an approach to reduce errors. In signal processing, a common approach is sampling Chebyshev nodes $x_{j} = \cos{\left(\frac{2j+1}{2n}\pi\right)}$, for $j \in [0, n]$. As for graph signal processing, it requires graph shifting as a mapping from eigenvalues of a normalized Laplacian into Chebyshev nodes. Designing such graph shift function is tough, since it desires eigendecomposition first, which the time complexity is $O(n^{3})$. However, through Chebyshev polynomial interpolation, we can bypass mapping eigenvalues while reducing errors.

\subsection{Chebyshev Polynomial Interpolation}
\begin{definition}[Chebyshev Polynomial Interpolation]
Given a target function $f(x)$ and a Chebyshev polynomial $p$ with $K$ orders, for $x \in [-1, 1]$, the target function $f(x)$ can be approximated as
\begin{equation}
\centering
\begin{split}
f(x) \approx p(x)
= \frac{1}{2}{\mu}_{0} + \sum_{k=1}^{\infty}{\mu}_{k}{T}_{k}(x)
\approx \frac{1}{2}{\mu}_{0} + \sum_{k=1}^{K}{\mu}_{k}{T}_{k}(x),
\end{split}
\end{equation}
where 
\begin{equation}
\centering
\begin{split}
\mu_{k} = \frac{2}{\pi}{\int}_{-1}^{1}\frac{f(x){T}_{k}(x)}{\sqrt{1-x^{2}}}\mathop{}\!\mathrm{d}{x} \approx \frac{2}{K+1}\sum_{j=0}^{K}f(x_{j})T_{k}(x_{j}),
\end{split}
\end{equation}
is the Chebyshev coefficients, and $x_{j}$ is a sampling Chebyshev node.
\end{definition}
Since the target graph frequency response function is unknown, we can replace the target eigenvalue $f(x_{j})$ with a learnable parameter $w_{j}$. It allows the model to simulate $f(x_{j})$ via gradient descent. Furthermore, we can simplify $\mu_{k}$ with a learnable parameter $w_{k}$. This is the original purpose proposed in ChebNet, where the Chebyshev coefficient vector $\bold{\mu} = [\mu_{0},\mu_{1},\cdots,\mu_{K}]$ is replaced by a learnable weight matrix $\mathbf{W}$.

Under the same order, if the target function is Dini–Lipschitz continuous, polynomial interpolation via Chebyshev basis converges faster than Bernstein~\citep{10.2307/2132695,Stein2003FourierAA}. If $x \in [-1,1]$, $C_{K}(f,x)$ denotes Chebyshev polynomial interpolation for target function $f(x)$ with $K$ orders. The convergence rate is $\lVert{C_{K}(f, x) - f(x)}\rVert_{\infty} \le -\frac{2}{\pi}\log(K^{-1})\omega(K^{-1})$, where $\omega(\cdot)$ denotes the modulus of continuity. Let $x \in [0, 1]$, $B_{K}(f,x)$ denotes Bernstein polynomial interpolation for target function $f(x)$ with $K$ orders, then the convergence rate is $\lVert{B_{K}(f, x) - f(x)}\rVert_{\infty} \le (1+\frac{1}{4}K^{-\frac{1}{2}})\omega(K^{-\frac{1}{2}})$. 

% With learnable damping factors, it is possible for a polynomial basis to simulate any polynomial basis. 

% \begin{lemma}
% Let $\lambda$ denotes the eigenvalues of the normalized Laplacian $\mathcal{L}$. Then, the eigenvalues of the renormalized Laplacian $\widetilde{\mathcal{L}}$ can be approximated as $\frac{\bar{d}}{\bar{d}+\eta}\lambda$ and the eigenvalues of the renormalized adjacency matrix $\widetilde{\mathcal{A}}$ can be approximated as $1 - \frac{\bar{d}}{\bar{d}+\eta}\lambda$, where $\bar{d}$ represents the average node degree and $\eta$ is the renormalization coefficient.
% \end{lemma}

\subsection{Gibbs Phenomenon and Gibbs Damping Factors}
\begin{table}[!t]
\centering
\caption{Performance (\%) comparison of ChebNet with or without various Gibbs damping factors in order $K=2$.}
\label{tab:chebnet_with_gibbs}
\begin{tabular}{l|cccccc}
\toprule
ChebNet &CoRA &CS &PM &Corn. &Texas &Wis. \\
\midrule
w/o G &78.39{\textpm}0.19 &68.43{\textpm}0.25 &67.77{\textpm}0.95 &67.57{\textpm}0.00 &77.57{\textpm}1.73 &72.94{\textpm}0.78\\
w/ J &75.86{\textpm}0.43 &68.37{\textpm}0.30 &73.98{\textpm}0.36 &70.27{\textpm}0.00 &83.24{\textpm}1.08 &84.31{\textpm}0.00\\
w/ La &73.08{\textpm}0.41 &67.38{\textpm}0.26 &75.60{\textpm}0.15 &70.27{\textpm}0.00 &81.89{\textpm}2.97 &84.31{\textpm}0.00\\
\bottomrule
\end{tabular}
\end{table}

\begin{table}[!t]
\begin{minipage}{0.48\linewidth}
\caption{Performance (\%) comparison of ChebNet with or without various Gibbs damping factors in CoRA with different order $K$.}
\label{tab:chebnet_with_gibbs_cora}
\centering
\begin{tabular}{|l|ccc|}
\toprule
K &w/o G &w/ J &w/ La \\
\midrule
4 &76.82{\textpm}0.29 &78.62{\textpm}0.44 &77.47{\textpm}0.48\\
6 &71.96{\textpm}0.51 &79.01{\textpm}0.42 &78.92{\textpm}0.45\\
8 &69.17{\textpm}0.74 &78.80{\textpm}0.32 &78.96{\textpm}0.38\\
\bottomrule
\end{tabular}
\end{minipage}
\begin{minipage}{0.48\linewidth}
\caption{Performance (\%) comparison of ChebNet with or without various Gibbs damping factors in CoRA with different order $K$.}
\label{tab:chebnet_with_gibbs_citeseer}
\centering
\begin{tabular}{|l|ccc|}
\toprule
K &w/o G &w/ J &w/ La \\
\midrule
4 &67.59{\textpm}0.65 &69.26{\textpm}0.25 &68.60{\textpm}0.23\\
6 &64.81{\textpm}0.39 &68.29{\textpm}0.43 &69.38{\textpm}0.20\\
8 &64.71{\textpm}0.49 &67.69{\textpm}0.25 &68.20{\textpm}0.29\\
\bottomrule
\end{tabular}
\end{minipage}
\end{table}

\begin{figure}[htbp]
\label{fig:gibbs_phenomenon}
\centering
\begin{minipage}[t]{0.48\textwidth}
\centering
\includegraphics[width=6cm]{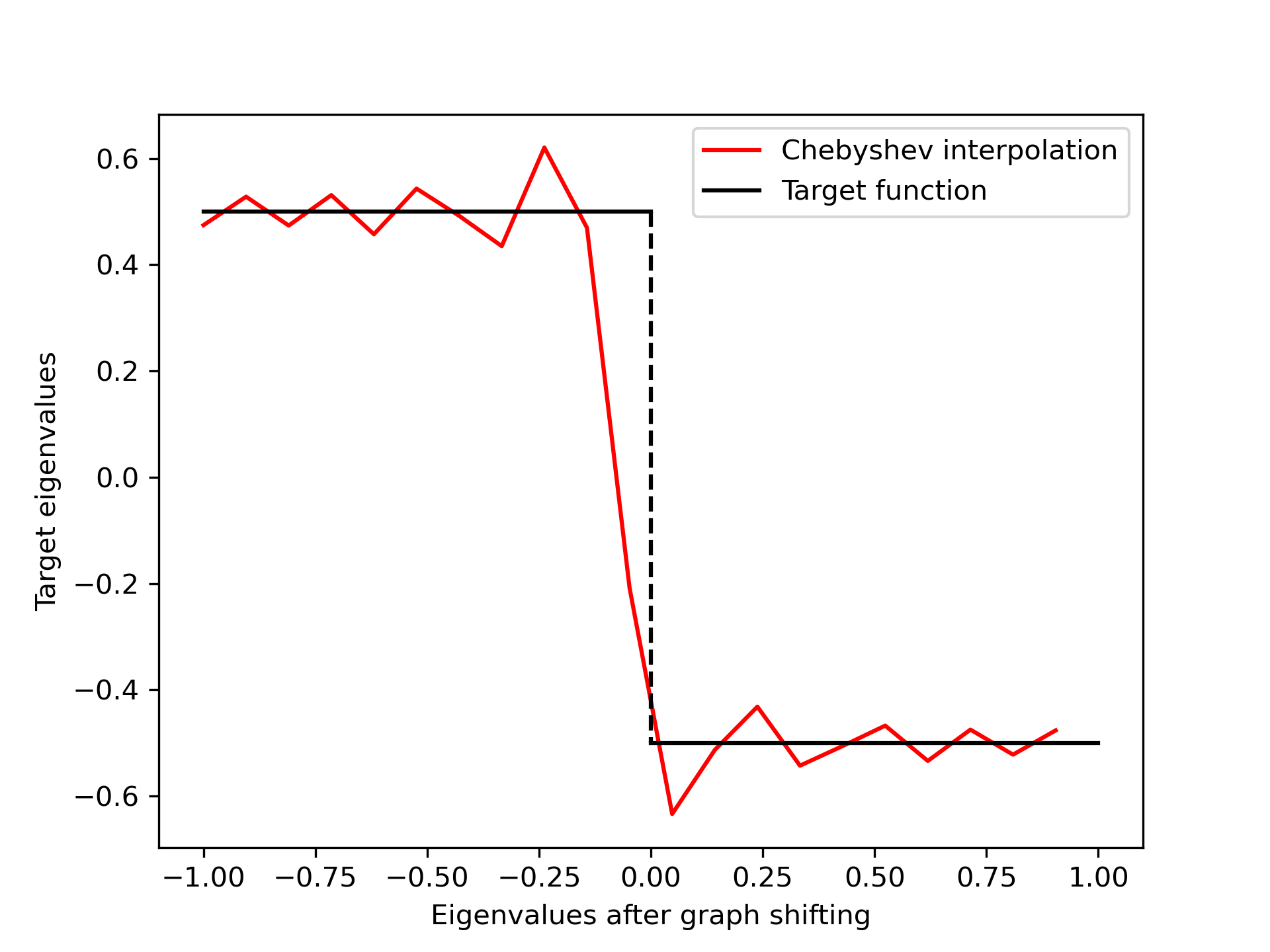}
\caption{The target function is discontinuous}
\end{minipage}
\begin{minipage}[t]{0.48\textwidth}
\centering
\includegraphics[width=6cm]{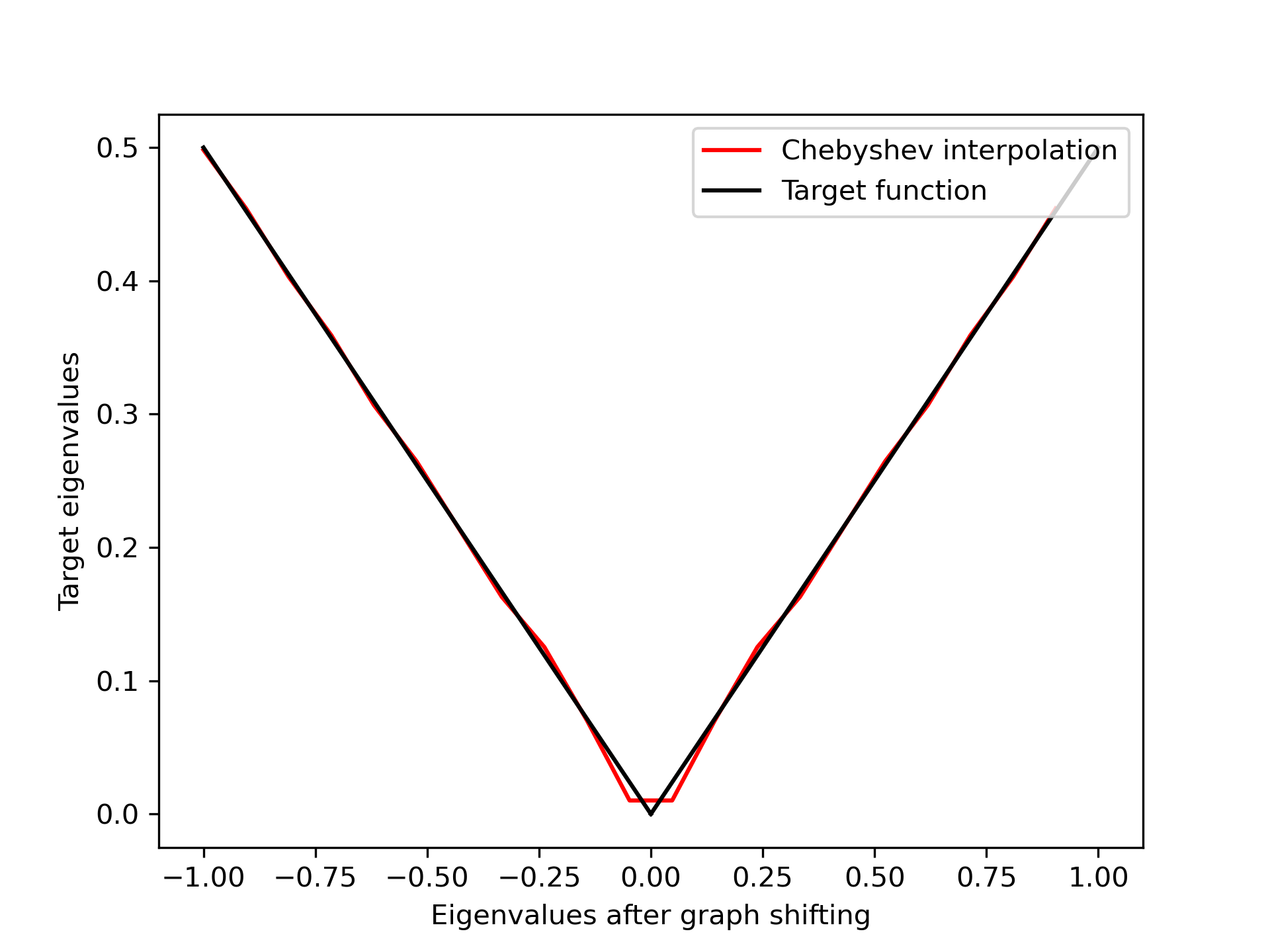}
\caption{The target function is singular}
\end{minipage}
\end{figure}

In real world datasets, the target graph frequency response function probably discontinuous or singular in the polynomial interpolation interval. It will cause intense oscillations near discontinuities or singularities during approximation process. This phenomenon is known as the Gibbs phenomenon. We demonstrate it in Figure~\ref{fig:gibbs_phenomenon}.

As some related research works~\citep{10.2307/2132695,Stein2003FourierAA} point out that when the Gibbs phenomenon occurs, the polynomial will not uniformly converge, but point-wisely converge except for discontinuities or singularities. It will substantially slow down the polynomial convergence rate, especially if there exists more than one discontinuities or singularities.

To mitigate Gibbs oscillations, we apply the Gibbs damping factor~\citep{RevModPhys.78.275} with each term of Chebyshev polynomials. It is a historical method that has been used in physics known as kernel polynomial method for more than two decades. To our best knowledge, it is the first time that Gibbs damping factors are adopted in graph convolutional networks. Gibbs damping factors are a family of coefficients that satisfy two conditions: (1)~$g_{0,K} = 1$. (2)~$\lim_{K \to \infty}{g_{1, K} \to 1}$. One common Gibbs damping factor is called the Jackson damping factor, which is defined as
\begin{equation}
\centering
\begin{split}
g_{k, K}^{\text{J}} = \frac{(K+2-k)\sin(\frac{\pi}{K+2})\cos(\frac{k\pi}{K+2})+\cos(\frac{\pi}{K+2})\sin(\frac{k\pi}{K+2})}{(K+2)\sin(\frac{\pi}{K+2})}.
\end{split}
\end{equation}
Another typical Gibbs damping factor is named Lanczos damping factor, which is defined as
\begin{equation}
\centering
\begin{split}
g_{k,K,m}^{\text{La}} = \left(\mathrm{sinc}(\frac{k}{K+1})\right)^{m}
= \left(\frac{\sin(\frac{k{\pi}}{K+1})}{\frac{k{\pi}}{K+1}}\right)^{m},\ \text{where}\ m \in \mathbb{Z}^{+}.
\end{split}
\end{equation}
When $m=3$, it is close to Jackson damping factor, but not strictly positive.
Thus, a graph convolution layer of ChebNet with Gibbs damping factors is defined as
\begin{equation}
\centering
\begin{split}
\mathbf{Z}^{(l+1)} = \sigma{(\sum_{k=0}^{K}{g}_{k,K}{T}_{k}(\widetilde{\mathcal{L}})\mathbf{Z}^{(l)}\mathbf{W}^{(l)})}.
\end{split}
\end{equation}
Table~\ref{tab:chebnet_with_gibbs} exhibits the ablation experimental results of ChebNet (when $K=2$) with and without various Gibbs damping factors for node classification on the citation networks~\citep{Sen_Namata_Bilgic_Getoor_Galligher_Eliassi-Rad_2008} and the webpage networks~\citep{lu2003link}. In Table~\ref{tab:chebnet_with_gibbs}, except in CoRA and CiteSeer, ChebNet with Gibbs damping factors exhibits better performance than without Gibbs damping factors. Table~\ref{tab:chebnet_with_gibbs_cora} and Table~\ref{tab:chebnet_with_gibbs_citeseer} illustrate Gibbs damping factors work for high order in CoRA and CiteSeer.

\begin{table*}[!t]
\centering
%\vspace{3mm}
\caption{Statistics of datasets}
\label{table:stat}
%\vspace{-2mm}
\begin{tabular}{lcccccccccc}
\toprule
\quad &CoRA &CS &PM &Corn. &Texas &Wis. &Film &Cham. &Squi. &WCS\\
\midrule
\#Nodes &2708 &3327 &19717 &183 &183 &251 &7600 &2277 &5201 &11701\\
\#Edges &10556 &9104 &88648 &298 &325 &515 &30019 &36101 &217073 &297110\\
\#Features &1433 &3703 &500 &1703 &1703 &1703 &932 &2325 &2089 &300\\
\#Classes &7 &6 &3 &5 &5 &5 &5 &5 &5 &10\\
$\mathcal{H}_{\mathrm{node}}(G)$ &0.83 &0.71 &0.79 &0.11 &0.07 &0.17 &0.16 &0.25 &0.22 &0.66\\
\bottomrule
\end{tabular}
\end{table*}

% \begin{table*}[!t]
% \centering
% \caption{Hyper-parameters of ChebJackNet}
% \label{table:hyper}
% %\vspace{-1mm}
% \begin{tabular}{lccccccc}
% \toprule
% Dataset     &$K$  &Learning rate  &$\ell^{2}$ regularization rate &Dropout rate   &Hidden dimension\\
% \midrule
% CoRA        &10      &$10^{-2}$    &$5\times10^{-4}$    &0.5      &64\\
% citepSeer    &10      &$10^{-2}$    &$5\times10^{-4}$    &0.5      &64\\
% PubMed      &10      &$10^{-2}$    &$5\times10^{-4}$    &0.5      &64\\
% Cornell     &10      &$10^{-2}$    &$5\times10^{-4}$    &0.5      &64\\
% Texas       &10      &$10^{-2}$    &$5\times10^{-4}$    &0.5      &64\\
% Wisconsin   &10      &$10^{-2}$    &$5\times10^{-4}$    &0.5      &64\\
% Film        &10      &$10^{-2}$    &$5\times10^{-4}$    &0.5      &64\\
% Chameleon   &10      &$10^{-2}$    &$5\times10^{-4}$    &0.5      &64\\
% Squirrel    &10      &$10^{-2}$    &$5\times10^{-4}$    &0.5      &64\\
% Wiki-CS     &10      &$10^{-2}$    &$5\times10^{-4}$    &0.5      &128\\
% \bottomrule
% \end{tabular}
% \vspace{-2mm}
% \end{table*}

\subsection{ChebGibbsNet and Analysis from Spectral Domain}
By restructuring ChebNet with learnable coefficients and Gibbs damping factors into the architecture summarized in Eq.~\ref{eq:SpecGCN}, we propose ChebGibbsNet.  
\begin{definition}[ChebGibbsNet]
\begin{equation}
\centering
\begin{split}
\hat{\mathbf{Y}}_{\text{ChebGibbsNet}} &=  \mathrm{Softmax}\left(\sum_{k=0}^{K}{w}_{k}{g}_{k,K}{T}_{k}(\mathbf{S})\cdot{f}_{\Theta}(\mathbf{X})\right),
\end{split}
\end{equation}
where ${w}_{k}$ is a learnable coefficient with one initialization that represents the corresponding Chebyshev coefficient $\mu_{k}$, $\mathbf{S} = \widetilde{\mathcal{A}}$ for homogeneous graphs and $\mathbf{S} = -\widetilde{\mathcal{A}}$ for heterogeneous graphs. Besides, the self-gated activation function SiLU~\citep{hendrycks2016gelu,elfwing2018silu,ramachandran2018swish} is employed in $f_{\Theta}(\mathbf{X})$ to stabilize performance.
\end{definition}

Recent research works~\citep{pmlr-v97-wu19e,nt2019revisiting,chien2021adaptive} indicate that a low-pass filter or band-stop filter can handle homogeneous graphs, and a high-pass filter or band-pass filter can handle heterogeneous graphs. Thus, the node homophily index $\mathcal{H}_{\text{node}}(G)$ is adopted in ChebGibbsNet to adaptively change the GSO. When $\mathcal{H}_{\text{node}}(G) \in (0.5, 1)$, it indicates that the graph is homogeneous. Then, $\widetilde{\mathcal{A}}$ as a GSO is adopted in the model. If learnable coefficients are non-negative, ChebGibbsNet with Gibbs damping factors performs low-pass or band-stop filtering on graph signals. When $\mathcal{H}_{\text{node}}(G) \in (0, 0.5)$, the graph filter of ChebGibbsNet with Gibbs damping factors becomes a high-pass or band-pass filter.

% \begin{theorem}[The stability of monomials based graph filters~\citep{9054072}]\label{theo:spf}
% Given a GSO $\mathbf{S}$ and a monomials based graph filter as $\mathbf{H}(\mathbf{S}) = \sum_{k=0}^{K}\zeta_{k}\mathbf{S}^{k}$ with $K \ge 2$ orders. Consider a perturbing GSO $\mathbf{S^{\prime}}$, then the following inequality is always holds:
% \begin{equation}
% \centering
% \begin{split}
% \lVert{\mathbf{H}(\mathbf{S}) - \mathbf{H}(\mathbf{S^{\prime}})}\rVert_{2} \le \frac{1}{4}\lVert{\boldsymbol{\zeta}_{-0}}\rVert_{1}(K^{2}-1)(\frac{K+1}{K-1})^{K}\lVert{\mathbf{S}-\mathbf{S^{\prime}}}\rVert_{2},
% \end{split}
% \end{equation}
% where $\boldsymbol{\zeta}_{-0} = [\zeta_{1}, \zeta_{2}, \cdots, \zeta_{K}]$ is the vector of damping factors for all terms except the $0$-th term.
% \end{theorem}
% Since Chebyshev-Jackson polynomials can be rewritten into monomials, ChebJackNet is stable for graph perturbation according to Theorem~\ref{theo:spf}.

There is an issue in graph representation learning called the over-smoothing issue~\citep{Li_Han_Wu_2018,Li_2019_ICCV}, which prevents deep graph convolutional networks from gaining better performance. To elaborate the over-smoothing issue, we need to analyze the global smoothness of a graph filter. Thus, we introduce the spectral gap~\citep{hoory2006expander} and the graph diffusion distance~\citep{MASUDA20171}.
\begin{definition}[Spectral gap]\label{def:specgap}
Given an undirected graph $G$, suppose that the eigenvalues of the normalized Laplacian $\mathcal{L}$ in ascending order are $\lambda_{0}, \lambda_{1}, \cdots, \lambda_{n-1}$. The spectral gap is defined as
\begin{equation}
\centering
\begin{split}
{\delta}_{j}(\lambda) = \lambda_{j} - \lambda_{j-1},\ \text{where}\ j \in [0, n-1].
\end{split}
\end{equation}
\end{definition}
The convergence rate of the stationary distribution of diffusion is positively related to the first spectral gap ${\delta}_{1}(\lambda)$~\citep{hoory2006expander}.

\begin{definition}[Graph diffusion distance]~\label{def:gdd}
Given an undirected graph $G$, the diffusion distance between node $u$ and node $v$ at discrete diffusion time $K$ is defined as
\begin{equation}
\centering
\begin{split}
s_{K}(u,v) = \sqrt{\sum_{w\in{\mathcal{V}}}\frac{\left(\mathbf{H}(\mathbf{S})(u,w)-\mathbf{H}(\mathbf{S})(v,w)\right)^{2}}{\pi{(w)}}},
\end{split}
\end{equation}
where $\pi{(w)}=\frac{\mathbf{D}(w,w)}{\sum_{(u,v)\in{\mathcal{E}}}\mathbf{D}(u,v)}$ is the stationary density at vertex $w$.
\end{definition}
With the increment of the discrete diffusion time, the graph diffusion distance is inevitably smaller. When the graph diffusion distance approaches $0$, the stationary distribution of diffusion is convergent. Then, the performance of a SpecGCN begins to decline. Therefore, from the view of the spectral gap and the diffusion distance, eliminating over-smoothing is tough. However, relieving or escaping from over-smoothing is relatively easy. Similar to the damping factor $(1-\alpha){\alpha}^{k}$ of PPR, Gibbs damping factor also will decay with the increase of the order of Chebyshev polynomials. With learnable coefficients, ChebGibbsNet can escape from over-smoothing as GPR-GNN and BernNet.
\section{Experiments}
% MLP
% ChebNet
% GCN
% SGC
% GAT
% APPNP
% JK-Net
% SSGC
% GPR-GNN
% BernNet
% p-GNN

\subsection{Datasets and Experimental Setup}
\noindent\textbf{Datasets.}
We use five different types graph datasets, three citation networks, three webpage networks, two Wikipedia networks, an actor co-occurrence network, and a Wikipedia computer science network, for node classification tasks. The detailed statistics of those datasets are shown in Table~\ref{table:stat}. Notice that a Wikipedia computer science network and citation networks are homogeneous graphs, and an actor co-occurrence network, Wikipedia networks and webpage networks are heterogeneous graphs.

\begin{table*}[t]
\centering
\vspace{-1mm}
\caption{Node classification accuracy (\%). Mean accuracy (\%) ± 95\% confidence interval. Boldface letters are used to mark the best results.}
%\vspace{-1mm}
\resizebox{1.01\columnwidth}{!}{
\begin{tabular}{lcccccccccc}
\toprule
Model &CoRA &CS &PM &Corn. &Texas &Wis. &Film &Cham. &Squi. &WCS\\
\midrule
MLP &60.33{\textpm}0.34 &59.97{\textpm}0.24 &73.49{\textpm}0.20 &71.89{\textpm}1.32 &79.73{\textpm}1.81 &\textbf{81.37{\textpm}0.98} &36.87{\textpm}0.21 &43.90{\textpm}0.64 &29.52{\textpm}0.64 &72.73{\textpm}0.21\\
ChebNet &79.74{\textpm}0.55 &69.42{\textpm}0.78 &69.20{\textpm}1.55 &69.46{\textpm}1.24 &79.46{\textpm}1.79 &76.08{\textpm}1.71 &34.56{\textpm}0.45 &45.70{\textpm}0.70 &32.48{\textpm}0.41 &76.70{\textpm}0.30\\
GCN &80.83{\textpm}0.44 &71.13{\textpm}0.18 &78.91{\textpm}0.18 &38.92{\textpm}2.16 &64.32{\textpm}2.02 &50.78{\textpm}2.23 &28.37{\textpm}0.55 &58.73{\textpm}0.46 &26.43{\textpm}0.14 &79.37{\textpm}0.12\\
SGC &77.90{\textpm}0.00 &\textbf{72.30{\textpm}0.00} &77.59{\textpm}0.03 &37.84{\textpm}0.00 &64.86{\textpm}0.00 &54.90{\textpm}0.00 &27.36{\textpm}0.10 &36.80{\textpm}0.00 &24.07{\textpm}0.10 &79.10{\textpm}0.12\\
APPNP &81.42{\textpm}0.12 & 71.74{\textpm}0.14 &79.60{\textpm}0.17 &70.27{\textpm}0.00 &75.68{\textpm}0.00 &80.39{\textpm}0.00 &36.12{\textpm}0.51 &44.45{\textpm}0.54 &29.69{\textpm}0.50 &72.60{\textpm}0.29\\
S$^{2}$GC &80.61{\textpm}0.05 &70.40{\textpm}0.00 &79.90{\textpm}0.00 &35.14{\textpm}0.00 &56.76{\textpm}0.00 &41.83{\textpm}0.92 &27.13{\textpm}0.45 &48.90{\textpm}0.00 &\textbf{35.93{\textpm}0.00} &77.75{\textpm}0.03\\
GPR-GNN &81.93{\textpm}1.08 &69.96{\textpm}0.58 &79.53{\textpm}0.25 &42.43{\textpm}6.63 &67.30{\textpm}3.51 &66.47{\textpm}3.33 &29.58{\textpm}1.70 &54.61{\textpm}2.42 &33.59{\textpm}2.25 &79.75{\textpm}0.43\\
BernNet &81.50{\textpm}0.41 &68.90{\textpm}1.48 &77.80{\textpm}0.37 &72.07{\textpm}1.27 &76.58{\textpm}1.27 &79.74{\textpm}0.92 &35.07{\textpm}0.44 &65.86{\textpm}0.72 &39.42{\textpm}1.60 &79.90{\textpm}0.17\\
\hline
\noalign{\vskip 0.5ex}
ChebGibbsNet &\textbf{82.42{\textpm}0.64} &71.14{\textpm}0.93 & \textbf{80.18{\textpm}0.65} &\textbf{78.11{\textpm}2.25} &\textbf{85.68{\textpm}2.43} &81.18{\textpm}1.80 &\textbf{37.90{\textpm}0.38} &\textbf{68.64{\textpm}1.04} &33.83{\textpm}0.48 &\textbf{80.19{\textpm}0.18}\\
\bottomrule
\end{tabular}
}
\label{table:res}
%\vspace{-3mm}
\end{table*}

\noindent\textbf{Citation Networks.}
CoRA, CiteSeer, and PubMed are standard citation network benchmark datasets. In these networks, every node represents a paper and every edge represents a citation from one paper to another. The edge direction is defined from a citing paper to a cited paper. The feature is a vocabulary of unique words. We follow~\citep{DBLP:conf/iclr/KipfW17} to preprocess the data into training, validation and test sets.

\noindent\textbf{Webpage Networks.}
It is a webpage dataset collected from computer science departments of various universities by Carnegie Mellon University~\citep{lu2003link}. In WebKB, there are 3 datasets named Cornell, Texas, and Wisconsin. Each node represents a webpage and each edge represents a hyperlink between two webpages. The edge direction is from the original webpage to the referenced webpage. The feature of each node is the bag-of-words representation of the corresponding page. For webpage networks, in order to evaluate supervised graph representation learning, we follow previous work~\citep{Pei2020Geom-GCN} to randomly split nodes of each class into 60\%, 20\%, and 20\% for training, validation, and test sets, respectively.

\noindent\textbf{Actor co-occurrence network.}
It is an actor-only induced subgraph of the film-director-actor-writer network. Each node represents an actor, and the edge between two nodes denotes co-occurrence documented on the same Wikipedia page. The feature of each node is the bag-of-words representation in the Wikipedia pages.

\noindent\textbf{Wikipedia networks.}
Chameleon and Squirrel are two page-page networks on specific topics in Wikipedia. In those datasets, nodes represent web pages and edges are mutual links between pages. Besides, node features correspond to several informative nouns in the Wikipedia pages. We classify the nodes into five categories in term of the number of the average monthly traffic of the web page.

\noindent\textbf{Wikipedia computer science network.}
Wikipedia computer science network was created by inspecting the list of 10000 prominent categories selected by the sanitizer and picking CS subject. The dataset consists of nodes corresponding to Computer Science articles, with
edges based on hyperlinks and 10 classes representing different branches of the field.

\noindent\textbf{Baselines, detailed setup and hyperparameters.}
To verify the superiority of our model, we introduce MLP, ChebNet, GCN, SGC, APPNP, S$^{2}$GC, GPR-GNN, and BernNet as baselines. For all these baselines, we use the default setting and parameters as described in the corresponding papers.

We train our models using an Adam optimizer~\citep{kingma2014adam} with a maximum of 10,00 epoch and early stopping with patience to be 30. Table~\ref{table:hyper} summaries other hyperparameters of ChebGibbsNet on all datasets. All experiments are tested on a Linux server equipped with an Intel i7-6700K 4.00 GHz CPU, 64 GB RAM, and an NVIDIA GeForce GTX 1080 Ti GPU. For a given setting, 10 experiments of different random seeds are conducted.

\begin{table*}[!t]
\centering
\caption{Hyper-parameters of ChebGibbsNet}
\label{table:hyper}
%\vspace{-1mm}
\begin{tabular}{lccccccc}
\toprule
Dataset     &$K$  &Learning rate  &$\ell^{2}$ regularization rate &Dropout rate   &Hidden dimension\\
\midrule
CoRA        &10      &$10^{-2}$    &$5\times10^{-4}$    &0.6      &64\\
CiteSeer    &10      &$10^{-2}$    &$5\times10^{-4}$    &0      &64\\
PubMed      &10      &$10^{-2}$    &$5\times10^{-4}$    &0.1      &64\\
Cornell     &10      &$10^{-2}$    &$5\times10^{-4}$    &0.2      &64\\
Texas       &10      &$10^{-2}$    &$5\times10^{-4}$    &0      &64\\
Wisconsin   &10      &$10^{-2}$    &$5\times10^{-4}$    &0.1      &64\\
Film        &10      &$10^{-3}$    &$5\times10^{-5}$    &0.6      &32\\
Chameleon   &10      &$10^{-2}$    &$5\times10^{-4}$    &0.5      &64\\
Squirrel    &10      &$10^{-2}$    &$5\times10^{-4}$    &0.3      &64\\
Wiki-CS     &10      &$10^{-2}$    &$5\times10^{-4}$    &0.2      &128\\
\bottomrule
\end{tabular}
\vspace{-2mm}
\end{table*}

\subsection{Experiment results and analysis}
Table~\ref{table:res} reports the mean of the node classification accuracy with standard deviation on the test set of each model. Except in CiteSeer, Wisconsin, and Squirrel, our ChebGibbsNet has a remarkable performance than other models.
\vspace{-3mm}
\section{Conclusion and Future Work}
In this research, we utilize the property of Chebyshev polynomials and Gibbs damping factors to propose ChebGibbsNet. We test our model in both homogeneous graphs and heterogeneous graphs. The experimental results demonstrate that ChebGibbsNet is widely applicable for realistic datasets. It verifies our assumption that Gibbs oscillations weaken the performance of ChebNet. Besides, our analysis for polynomial interpolation in graph signal processing illustrates that Chebyshev polynomials have a faster convergence rate than Bernstein polynomials. Out of curiosity, we will tap potentials of other orthogonal polynomials for exploiting the future of graph representation learning.



\bibliography{iclr2023_conference}
\bibliographystyle{iclr2023_conference}

\end{document}